%% file: Formatting-Instructions-LaTeX-2026.tex
\documentclass[letterpaper]{article} 
\usepackage{aaai2026}  
\usepackage{times}  
\usepackage{helvet}  
\usepackage{courier}  
\usepackage[hyphens]{url}  
\usepackage{graphicx} 
\usepackage{natbib}  
\usepackage{caption} 
\usepackage{algorithm}
\usepackage{algorithmic}	
\usepackage{amsmath}	
\usepackage{amssymb}	
\usepackage{booktabs}
\usepackage{microtype}
\usepackage{epsfig}
\usepackage{placeins}
\usepackage{pifont}
\newcommand{\XSolidBrush}{\ding{55}}
\newcommand{\Checkmark}{\ding{51}}
\usepackage{enumitem}
\usepackage{tabularx}
\usepackage{xstring}
\usepackage{multirow}
\usepackage{xspace}
\usepackage{url}
\usepackage{subcaption}
\usepackage{xcolor}
\usepackage[hang,flushmargin]{footmisc}
\usepackage{colortbl}
\usepackage[table,xcdraw]{xcolor}

\usepackage{newfloat}
\usepackage{listings}
\DeclareCaptionStyle{ruled}{labelfont=normalfont,labelsep=colon,strut=off} 
\floatstyle{ruled}
\newfloat{listing}{tb}{lst}{}
\floatname{listing}{Listing}
\title{Zero-Shot Open-Vocabulary Human Motion Grounding with Test-Time Training}
\author{
    Yunjiao Zhou, Xinyan Chen, Junlang Qian,
   Lihua Xie, Jianfei Yang\textsuperscript{\* }\thanks{Corresponding author.}\\
}
\affiliations{
    Nanyang Technological University, Singapore\\

    \{yunjiao001, chen1909, junlang001\}@e.ntu.edu.sg, \{elhxie, jianfei.yang\}@ntu.edu.sg
}

\usepackage{bibentry}

\begin{document}

\maketitle

\input{chapter/00_abstract}

\begin{links}
    \link{Code}{https://github.com/pridy999/ZOMG}
\end{links}

\input{chapter/01_introduction}

\input{chapter/02_related}
\input{chapter/03_method_compress}
\input{chapter/04_experiment}
\input{chapter/10_conclusion}

\section{Acknowledgments}
This work is supported by Ministry of Education, Singapore, under AcRF TIER 1 Grant RG64/23 and National Research Foundation of Singapore Medium-sized Centre for Advanced Robotics Technology Innovation. This work is jointly supported by MOE Singapore Tier 1 Grant RG83/25, RS36/24 and a Start-up Grant from Nanyang Technological University.

\bibliography{aaai2026}

\end{document}

%% file: chapter/00_abstract.tex
\begin{abstract}

Understanding complex human activities demands the ability to decompose motion into fine-grained, semantic-aligned sub-actions. This motion grounding process is crucial for behavior analysis, embodied AI and virtual reality.
Yet, most existing methods rely on dense supervision with predefined action classes, which are infeasible in open-vocabulary, real-world settings. In this paper, we propose ZOMG, a zero-shot, open-vocabulary framework that segments motion sequences into semantically meaningful sub-actions without requiring any annotations or fine-tuning. Technically, ZOMG integrates (1) language semantic partition, which leverages large language models to decompose instructions into ordered sub-action units, and (2) soft masking optimization, which learns instance-specific temporal masks to focus on frames critical to sub-actions, while maintaining intra-segment continuity and enforcing inter-segment separation, all without altering the pretrained encoder.
Experiments on three motion-language datasets demonstrate state-of-the-art effectiveness and efficiency of motion grounding performance, outperforming prior methods by +8.7\% mAP on HumanML3D benchmark. Meanwhile, significant improvements also exist in downstream retrieval, establishing a new paradigm for annotation-free motion understanding.

\end{abstract}

%% file: chapter/01_introduction.tex
\section{Introduction}

Understanding human motion at a fine-grained level is vital for tasks such as behavior analysis, embodied AI, and virtual reality~\cite{zhang2023finemogen, zhou2024avatargpt, zhou2023metafi++, yang2022metafi}. However, real-world motion is temporally unstructured and composed of overlapping sub-actions without explicit boundaries. This absence of temporal modularity hinders downstream applications such as motion retrieval~\cite{xue2025shotvl, zhou2024adapose} and generation~\cite{zeng2025light}, which rely on semantically meaningful motion units for reasoning and interaction. In such cases, motion grounding, which decomposes continuous motion streams into semantically coherent units, plays a pivotal role in enabling structured motion representations and flexible motion interaction.


Existing motion grounding methods~\cite{chen2021end, yang2022tubedetr} typically rely on closed-set assumptions~\cite{zhou2023tent, radford2021learning}, training models to align motion sequences with labels from a fixed action vocabulary (e.g., ``walk", ``wave"). However, real-world motion is inherently open-ended and compositional, with free-form expressions describing subtle, overlapping sub-actions (e.g., ``a person sits down while waving”) that cannot be captured by predefined atomic labels. This rigid setup fundamentally limits scalability and generalization, highlighting the need for open-vocabulary grounding that aligns motion with natural language. 
In addition, the scarcity of fine-grained annotations and the diversity of real-world expressions motivate a zero-shot setting~\cite{punnakkal2021babel, qian2025beyond}, where models are required to generalize to novel queries without task-specific supervision. 
Therefore, our research focus is \textbf{\textit{to achieve reliable and efficient motion grounding under a zero-shot open-vocabulary setting.}}

\input{figs/top}

However, realizing zero-shot open-vocabulary motion grounding poses several challenges. 
First, open-world descriptions are free-form and structurally ambiguous, often lacking explicit cues regarding the number of sub-actions and their complex temporal interactions~\cite{liu2022exploring}. For instance, phrases like ``walks confidently and waving” mix concurrent and sequential elements, making it difficult for rule-based models~\cite{wu2022rule} to extract correct segments.
Second, while existing motion-language models~\cite{tevet2022motionclip} capture global semantics, grounding requires finer temporal resolution to detect subtle transitions between sub-actions. This necessitates frame-level reasoning, which current sequence-level encoders are not inherently equipped to handle.
Third, even when frame-level features are available, motion exhibits strong temporal continuity and entanglement. Unlike static images, individual frames often lack standalone semantic identity, and sub-actions emerge only through inter-frame dynamics. Without modeling these dependencies, frame-wise analysis alone is insufficient for accurate grounding.


Given these challenges, it is necessary to obtain finer-grained representations for both text and motion. On the language side, we compensate for the absence of segment-level annotations by using large language models (LLMs)~\cite{zhao2023survey} to decompose free-form descriptions into ordered sub-action units. These unit-level queries provide semantic guidance for open-vocabulary matching. On the motion side, directly supervising frame-level alignment is impractical due to annotation scarcity. 
Pretrained motion-language models, despite being trained only at the sequence level, implicitly encode the inter-frame transitions necessary for grounding~\cite{wang2023glanet}.
This suggests that exploiting such temporal structure already embedded in the representation space is promising for fine-grained motion grounding. To make this structure accessible without altering model parameters, we design a test-time training approach. It introduces a small set of variables for each input, which are optimized to selectively attend to sub-action-relevant frames. This adaptation refines the attention behavior of pretrained model~\cite{lee2019self} on a per-instance basis, allowing precise localization of motion segments while maintaining generalization in zero-shot settings.

In this work, we propose ZOMG, a test-time grounding framework that extracts fine-grained motion structure from pretrained motion-language models without annotations.
ZOMG bypasses the need for fine-tuning by introducing a lightweight test-time training stage that transforms sequence-level representations into temporally grounded sub-actions.
It consists of two modules:
(1) Language Semantic Partition (LSP) uses an LLM to decompose free-form textual descriptions into semantically coherent and temporally ordered sub-action queries, serving as anchors for subsequent grounding;
(2) Soft Masking Optimization (SMO) aligns each query with motion by optimizing frame-wise soft masks, guided by semantic alignment and structural regularizations to ensure segment separability and continuity.
Experiments across large-scale datasets demonstrate that ZOMG improves grounding accuracy by up to +8.7\% mAP on HumanML3D and significantly boosts downstream motion-text retrieval. Despite these gains, it remains highly efficient, requiring only 0.5K optimization parameters and achieving over 3× inference speedup compared to existing TTT methods, enabling practical annotation-free deployment.

The contributions are summarized as follows:
\begin{itemize}

    \item We introduce ZOMG, the first framework that enables annotation-free and open-vocabulary motion grounding, achieving +8.7\% mAP improvement on HumanML3D.

    \item We propose a novel test-time training scheme combining LLM-guided decomposition and soft masking optimization, uncovering the fine-grained temporal structure from pretrained motion-language models.

    \item Extensive experiments show that ZOMG achieves state-of-the-art grounding accuracy and significantly improves motion-text retrieval. Despite these gains, it remains highly efficient at test time, supporting its practical deployment in real-world settings.

\end{itemize}

%% file: figs/top.tex
\begin{figure}[tp]
    \centering
    \includegraphics[width=\linewidth]{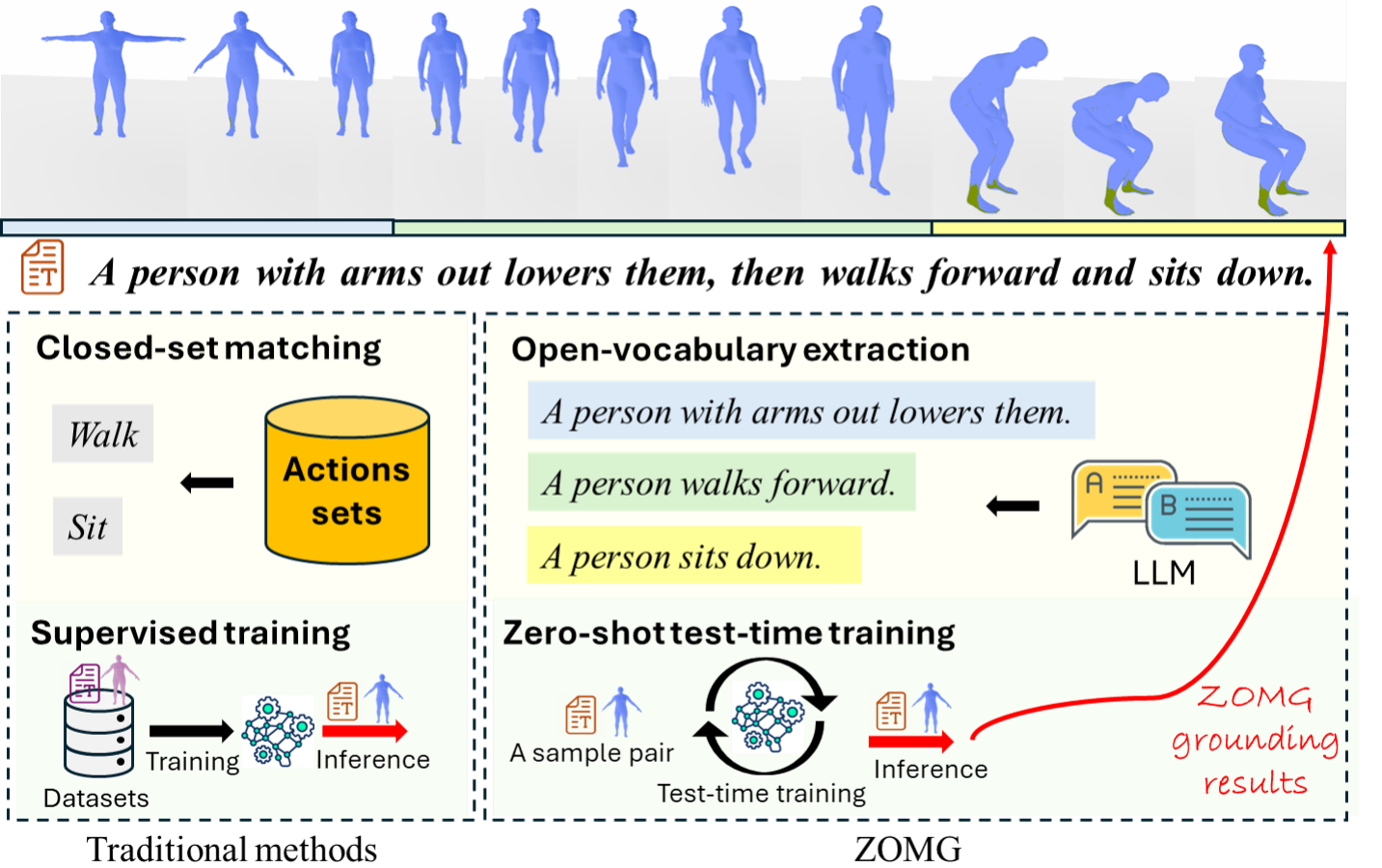}
    
    \caption{Motion grounding illustration of ZOMG.}
    \label{fig:top}
\end{figure}

%% file: chapter/02_related.tex
\section{Related Work}
\label{sec:related}



\subsection{Temporal Grounding}
Temporal grounding seeks to localize video segments that correspond to free-form textual descriptions~\cite{chao2018rethinking, zhao2017temporal, yan2023unloc, nguyen2025multi}. Early approaches like TALL~\cite{gao2017tall} and 2D-TAN~\cite{zhang2020learning} formulate this task as moment retrieval via cross-modal attention and sliding-window proposals, trained under dense annotations and closed-set vocabularies.
Recent work has explored open-vocabulary and zero-shot settings by leveraging pretrained vision-language models. For example, T3AL~\cite{liberatori2024test} uses CLIP-based retrieval to match language queries with video frames, while STALE~\cite{nag2022zero} introduces a one-stage model with parallel streams for localization and classification. X-POOL~\cite{gorti2022x} and SeViLA~\cite{yu2023self} further enhance grounding via frozen VLMs~\cite{xu2021vlm, bordes2024introduction}, augmented with adapters or prompting strategies.
These methods showcase the power of pretrained representations for temporal localization, but are largely limited to video domains, where large-scale vision-language models provide strong priors. In contrast, motion data lacks such pretrained encoders, making zero-shot grounding in this space substantially more challenging.



\subsection{Motion Understanding}

Understanding and generating human motion from language has drawn increasing interest for applications in animation, robotics, and embodied AI. Sequence-level models like TEMOS~\cite{petrovich2022temos}, TMR~\cite{petrovich2023tmr}, and TEACH~\cite{athanasiou2022teach} learn joint motion-text representations via contrastive or generative training. Others, including MotionCLIP~\cite{tevet2022motionclip} and Motion-X~\cite{lin2023motion}, incorporate CLIP priors to encode human motion, but operate at the sequence level, limiting their capacity for segment-level reasoning.
Recent studies have begun to explore finer temporal structure through action segmentation~\cite{wang2023actionclip, kong2019mmact, punnakkal2021babel} and fine-grained motion retrieval~\cite{li2024motion, wang2024text, zhang2023finemogen}. However, these methods typically depend on predefined taxonomies or task-specific labels, and seldom address frame-level grounding in open settings.
In contrast, our work aims to understand fine-grained motion under open-vocabulary, zero-shot conditions. 
Without predefined categories, our method extracts semantically coherent motion units that can be reused across diverse tasks, enhancing both generalization and interpretability.

%% file: chapter/03_method_compress.tex
\section{Method}
\label{sec:method}
\input{figs/framework}

\subsection{Problem Definition}

Zero-shot open-vocabulary human motion grounding aims to segment compositional motion sequences into semantically coherent units aligned with free-form text, without predefined action classes or temporal annotations. Formally, given a text \( \mathcal{T} \) and a motion sequence \( \mathcal{M} = \{ m_1, \dots, m_L \} \) of $L$ frames, the goal is to segment $\mathcal{M}$ into \( \{ S_1, \dots, S_k \} \), where each segment $S_i$ corresponds to a sub-action aligned with a span  \( \mathcal{T}_i \subseteq \mathcal{T} \).
Our framework, shown in Fig.~\ref{fig:template}, includes: (1) Motion-Language Pretraining, which establishes a fundamental understanding of global motion-text semantic alignment, and (2) Test-Time Grounding, which adapts the pretrained model on a per-instance basis to segment motion sequences in alignment with the input text.

\subsection{Motion-Language Pretraining}

Motion-language pretraining focuses on sequence-level alignment, implicitly capturing inter-frame temporal dynamics to provide cues for downstream segmentation even without segment-level supervision.
Given paired motion-text samples $(\mathcal{M}, \mathcal{T})$, the motion encoder $E_M$ maps $\mathcal{M}$ to per-frame features $F = \{f_1, \dots, f_L\}, f_i  \in \mathbb{R}^d $, while the text encoder $E_T$ produces a global text embedding $t \in \mathbb{R}^d$. To retain fine-grained temporal cues, $E_M$ explicitly computes these frame-level representations before any temporal aggregation. An attention pooling module $P(\cdot)$ then computes the motion embedding $m \in \mathbb{R}^d$ via learned weights, integrating local semantics with temporal salience.
The model is trained with a contrastive loss~\cite{radford2021learning}:
\begin{equation}
   \mathcal{L}_{c}=-\log \frac{\exp(m 
\cdot t^+/\tau)}{\sum_{t \in \{t^+, t^-\}}\exp(m \cdot t/\tau)},
\end{equation}
where $t^+$ and $t^-$ denote positive and negative text samples, respectively.
This pretraining yields general-purpose encoders that implicitly capture motion structure, laying the foundation for zero-shot fine-grained grounding.

\subsection{Test-Time Grounding}
While existing approaches often rely on fine-tuning~\cite{wortsman2022robust} or adapter-based training~\cite{wang2020k} to specialize pretrained models for grounding tasks, such methods require nontrivial amounts of segment-level supervision and may overfit to training distributions. This contradicts our goal of fully zero-shot, open-vocabulary motion grounding.
To address this, we adopt a test-time training strategy~\cite{sun2020test, liu2021ttt++}, which enables instance-specific adaptation without modifying the pretrained model. It introduces:  
(1) LSP, which leverages LLMs to decompose free-form textual descriptions into temporally ordered sub-action units, providing structured semantic anchors for grounding.  
(2) SMO, which performs instance-specific optimization to generate frame-wise soft masks, identifying sub-action-relevant frames without supervision.

\subsubsection{Language Semantic Partition}
Zero-shot motion grounding lacks segment-level annotations, making it difficult to localize complex instructions over time. Traditional rule-based or syntactic parsing struggles with ambiguity and compositionality in open-vocabulary descriptions~\cite{fitzgerald2018large}. 
To bridge this gap, we harness LLMs' reasoning to decompose complex descriptions into coherent, temporally ordered sub-action units. It captures discourse-level semantics, accurately extracting motion-relevant units from abstract texts. 
To ensure reliable decomposition, we impose two criteria on sub-actions:  
(1) \textit{Semantic completeness}, ensuring each unit forms a coherent motion;  
(2) \textit{Temporal decomposability}, preserving the logical order of actions.  
These criteria are enforced through a structured prompt with in-context examples that guide the LLM to decompose the free-form texts. 
To enhance robustness, we apply LLM-based paraphrasing followed by majority voting to produce the final decomposition:
\[
\mathcal{T} \rightarrow \{\mathcal{T}_1, \mathcal{T}_2, ..., \mathcal{T}_k\}
\]
Further implementation details, including prompt design and voting strategies, are provided in Appendix~A.2.

\subsubsection{Soft Masking Optimization}


The second stage refines the coarse, sequence-level alignment learned during pretraining into a fine-grained, temporally localized understanding of motion. While attention pooling aggregates frame-wise features into global embeddings, it operates with a fixed receptive field over the full sequence, often suppressing local variations and blurring sub-action boundaries. This becomes particularly limiting in zero-shot settings, where unseen actions must be inferred from latent structure without supervision.
To address this, SMO introduces a learnable soft mask that adapts attention pooling to each sub-action query. By dynamically reweighting frame contributions, the mask enables the model to focus on semantically relevant segments while preserving broader temporal context.

Given a set of $k$ sub-action queries $\{ \mathcal{T}_1, \dots, \mathcal{T}_k \}$, each query $\mathcal{T}_i$ is encoded as a semantic embedding $\hat{t}_i \in \mathbb{R}^d$ using the frozen $E_T$. The motion sequence $\mathcal{M} = \{ m_1, \dots, m_L \}$ is encoded by $E_M$ into frame-wise features $F = \{ f_1, \dots, f_L \}$.
To identify sub-action-relevant frames, we introduce a set of learnable soft masks $[M] = \{ M_1, \dots, M_k \}$, where each $M_i \in \mathbb{R}^L$ assigns a relevance score to every frame for the $i$-th sub-action. Since each frame may be semantically relevant to multiple sub-actions, we normalize the masks across sub-actions at each frame via softmax~\cite{liu2016large}:
\begin{equation}
\hat{M}_{i,t} = \frac{\exp(M_{i,t})}{\sum_{j=1}^{k} \exp(M_{j,t})},
\end{equation}
where $\hat{M}_{i,t} \in (0, 1)$ represents the normalized probability that frame $t$ semantically belongs to sub-action $T_i$. This normalization avoids degenerate solutions (e.g., all masks attending to all frames), encourages competition between masks, and promotes segment-level disentanglement.

Each normalized soft mask $\hat{M}_i \in \mathbb{R}^L$ is used to reweigh frame-wise features via element-wise multiplication:
\begin{equation}
\hat{F}_i = \hat{M}_i \cdot F = \{ \hat{M}_{i,1} f_1, \hat{M}_{i,2} f_2, \dots, \hat{M}_{i,L} f_L \}.
\end{equation}
This operation serves as a soft selection over time, allowing the model to emphasize semantically important frames while retaining the full temporal resolution.

Unlike hard windowing or static pooling, soft masking permits continuous gradients and supports query-specific focus without requiring discrete decisions. The resulting features $\hat{F}_i$ are then aggregated via attention pooling to produce the sub-action embedding:
\begin{equation}
\hat{m}_i = P(\hat{F}_i) \approx P([f_t]), \quad t \in T_i.
\end{equation}
Here, $\hat{M}_i$ acts as a semantic pre-filter that suppresses irrelevant frames by scaling down their feature amplitudes before pooling. This reshapes the pooling’s receptive field around sub-action-relevant regions, while preserving the inter-frame dynamics learned during pretraining.

To supervise the learning of query-specific soft masks, we adopt the same contrastive loss as pretraining, encouraging each sub-action’s motion embedding $\hat{m}_i$ to align with its corresponding text embedding $\hat{t}_i$. Formally:
\begin{equation}
    \mathcal{\hat{L}}_{c}=-\log \frac{\exp(\hat{m}_i 
\cdot \hat{t}_i/\tau)}{\sum_{\hat{t} \in \{\hat{t}_i, \hat{t}^-\}} \exp(\hat{m}_i \cdot \hat{t}/\tau)},
\end{equation}
where $\hat{t}^-$ are negative sub-action embeddings from the same sequence. Unlike pretraining where negatives are drawn from different samples, here we emphasize intra-sample sub-action discrimination, promoting segment-level orthogonality.



\textbf{Mask constrains.} Although soft masking enables query-specific frame reweighting, learning high-quality masks remains under-constrained. Without explicit supervision, the model may produce ambiguous or unstable masks that hinder precise localization. In particular, effective sub-action masks should satisfy two critical properties: (1) \textit{inter-segment separability}, ensuring that different sub-actions attend to distinct temporal regions; and (2) \textit{intra-segment continuity}, preserving the temporal coherence within each segment. These properties are crucial for disentangling overlapping motion patterns and avoiding fragmented attention.

To promote inter-segment separability, we introduce an exclusivity loss $\mathcal{\hat{L}}_e$ that penalizes frame-level overlap between soft masks of different sub-actions, promoting temporal separation across sub-actions:
\begin{equation}
\mathcal{\hat{L}}_e = \frac{1}{k(k-1)} \sum_{i \ne j} \hat{M}_i^\top \hat{M}_j,
\end{equation}
where $\hat{M}_i^\top \hat{M}_j$ measures how much two sub-actions attend to the same frames. This constraint encourages orthogonality across masks, helping to disentangle temporally overlapping motion patterns and sharpen segment boundaries.

For intra-segment continuity, we impose a smoothness loss $\mathcal{\hat{L}}_s$ that penalizes abrupt transitions in mask values across consecutive frames:
\begin{equation}
\mathcal{\hat{L}}_s = \frac{1}{k(L-1)} \sum_{i=1}^{k} \sum_{t=1}^{L-1} (\hat{M}_{i,t+1} - \hat{M}_{i,t})^2.
\end{equation}
reflecting the temporal consistency of human motion. This regularization enforces local consistency within each mask, suppressing fragmented or unstable masks and encouraging the formation of coherent, contiguous segments.

Together, the two constraints form a complementary design that enforces both segment-level exclusivity and intra-segment continuity, enforcing meaningful structural priors over the learned soft masks.
The final test-time objective integrates alignment supervision with both constraints:
\begin{equation}
\mathcal{\hat{L}} = \alpha \cdot \mathcal{\hat{L}}_c + \beta \cdot \mathcal{\hat{L}}_e + \gamma \cdot \mathcal{\hat{L}}_s,
\end{equation}
where $\alpha, \beta, \gamma$ control the trade-off between semantic alignment, segment separability, and temporal coherence.

\textbf{Mask-Based Segment Decoding.} 
After obtaining the optimized soft masks, we convert them into discrete temporal segments by assigning each frame to the sub-action with the highest activation. Specifically, for each frame $t$, we compute 
\begin{equation}
    y_t = \arg\max_i \hat{M}_{i,t},
\end{equation}
resulting in a frame-level label sequence ${y_1, \dots, y_L}$. Contiguous spans with identical labels are then grouped into segments $\{S_1, \dots, S_k\}$.
Unlike threshold-based methods that require hyperparameter tuning, our decoding is parameter-free and directly reflects model confidence, yielding semantically aligned and temporally coherent segmentation.

%% file: figs/framework.tex
\begin{figure*}[tp]
    \centering
    \includegraphics[width=1\linewidth]{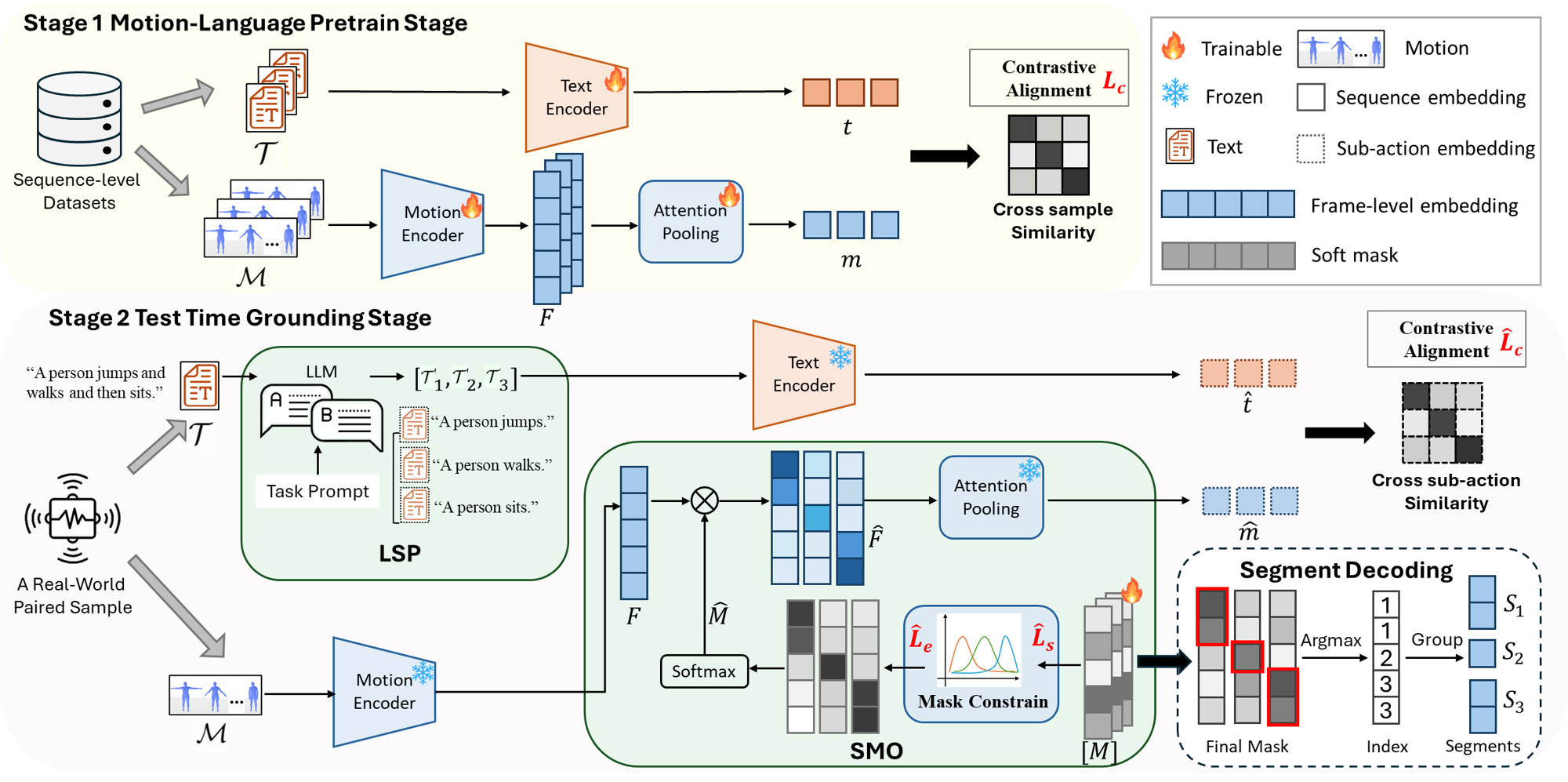}
    
    \caption{Overall framework of ZOMG.}
    \label{fig:template}
\end{figure*}

%% file: chapter/04_experiment.tex
\section{Experiment}
\input{tables/grounding}
\input{figs/tsne}
\input{figs/combo}
\input{figs/visual}
\subsection{Setup}
\subsubsection{Dataset.} 
We evaluate ZOMG on three public motion-language datasets: HumanML3D~\cite{guo2022generating} (29,232 motion-text pairs), KIT-ML~\cite{plappert2016kit} (3,911 pairs), and BABEL~\cite{punnakkal2021babel}, which includes dense annotations over 100 hours. For grounding, we select samples containing multiple sub-actions identified by the LLM, yielding 11,220 (HumanML3D), 1,207 (KIT-ML), and 2,210 (BABEL) test instances.


\subsubsection{Baseline.} We adopt MotionCLIP~\cite{tevet2022motionclip} as a baseline directly computing similarity between motion and text embeddings. 
Due to the lack of zero-shot grounding baselines in the motion domain, we compare with SOTA video grounding methods, including non-adaptive models (TCAM~\cite{belharbi2023tcam}, FreeZAD~\cite{han2025training}) and test-time training (TTT) methods (T3AL~\cite{liberatori2024test}, AdaZAD~\cite{han2025training}).
We evaluate grounding quality using mean Average Precision (mAP) computed over multiple temporal Intersection-over-Union (IoU) thresholds. Following standard practice, we report mAP averaged over thresholds of [0.3: 0.1: 0.8] on all three datasets.



\subsubsection{Implementation.} We adopt transformer-based encoders for motion and text. 
Motion is represented in H3D format with 251/263-dimensional features per frame~\cite{lu2023humantomato}.
We use Qwen-Plus~\cite{yang2025qwen3} LLM for sub-action decomposition and optimize soft masks for each query 100 steps. Loss weights for $\alpha, \beta, \gamma$ are set at 1, 0.005 and 100, respectively. All experiments are conducted on a single NVIDIA RTX 3090 GPU using PyTorch with mixed precision.


\subsection{Motion Grounding Performance}
\subsubsection{Quantative Results.}
Shown in 
Table~\ref{tab:grounding},
non-adaptive methods directly apply pretrained knowledge without instance-level adaptation, leading to sub-optimal performance due to their inability to capture fine-grained motion-language alignment.
In contrast, test-time tuning approaches significantly improve grounding quality by optimizing lightweight parameters for each instance.
Among them, ZOMG achieves the best performance, surpassing the previous SOTA method AdaZAD by +8.69\%, +9.62\%, and +3.29\% mAP on HumanML3D, KIT-ML, and BABEL, respectively. These consistent gains validate the robustness and effectiveness of ZOMG in producing fine-grained, semantically aligned motion segments under zero-shot open-vocabulary conditions.

\subsubsection{Visualization Results.}
\input{tables/ablation}
Figure~\ref{fig:combo} presents qualitative comparisons across different methods. While baselines often produce fragmented or misaligned segments, ZOMG generates temporally coherent and semantically meaningful outputs that align well with the intended sub-actions. Notably, the predicted segments of ZOMG closely match the optimized mask heatmaps in Figure~\ref{fig:box} (top), confirming that soft masking effectively captures subtle temporal structure and aligns motion with nuanced open-vocabulary semantics.  Additional visualizations are provided in the Appendix~A.4.

\subsubsection{Semantic Alignment Evaluation.}
To further assess grounding quality, we evaluate the semantic alignment between the segmented motion clips and their corresponding texts. After applying different methods to segment the motion into sub-actions, we compute the motion-text similarity for each pair using a pretrained motion-language encoder. This metric serves as an intuitive proxy for alignment quality: higher similarity indicates that the segmented motion more faithfully reflects the intended semantics.  
As shown in Figure~\ref{fig:visual}, ZOMG consistently achieves stronger motion-text alignment, reflected both in its top-ranking similarity scores and in a distribution skewed toward higher values.
This alignment-based evaluation highlights ZOMG’s ability to produce semantically faithful segments, offering clear advantages in real-world, open-vocabulary scenarios.

\subsection{Ablation and Analytical Studies}
\subsubsection{Component Ablation.}
Table~\ref{tab:ablation} assess the contribution of each component in ZOMG’s test-time grounding stage on HumanML3D.
Without test-time training, the non-adaptive baseline performs poorly (5.86\% mAP), highlighting the need for instance-specific optimization.
Incorporating LSP yields clear gains over rule-based segmentation, demonstrating the value of LLM-guided sub-action decomposition.
To evaluate SMO, we progressively activate its components.
Introducing the soft mask [$M$] significantly boosts performance, suggesting that optimizing frame-wise weights is critical for aligning motion with language.
Adding $\hat{\mathcal{L}}_s$ or $\hat{\mathcal{L}}_e$ further improves results, with the full model achieving the best performance (50.46\% mAP).
These findings confirm the complementary roles of the mask constraints and the importance of each module for effective zero-shot grounding.

\input{tables/efficiency}

\input{tables/aug_retrival_compress_hml3d}

\subsubsection{Mask Quality Analysis.}
To evaluate the learned masks, we examine two key properties: inter-segment separation and intra-segment continuity, both essential for discovering temporally distinct and semantically coherent sub-actions.
\textbf{Qualitatively}, Figure~\ref{fig:box} (top) shows that the optimized masks activate over localized regions with smooth transitions and minimal overlap. Each mask captures a distinct sub-action with clear and interpretable boundaries, dynamically adjusting the receptive field of pooling layers to focus on the most relevant frames.
\textbf{Quantitatively}, We further assess mask quality using grounding error and continuity rate over TTT iterations (Figure~\ref{fig:line}). In (a), ZOMG consistently reduces grounding error across iterations and outperforms all baselines. Notably, this performance improvement closely follows the decline of the exclusivity loss $\hat{\mathcal{L}}_e$, indicating that enhanced inter-mask separation directly benefits grounding accuracy. In (b), continuity rate quickly declines and remains low, reflecting the effectiveness of the smoothness constraint $\hat{\mathcal{L}}_s$. Together, these results confirm that ZOMG learns structured and disentangled masks through efficient test-time adaptation.

\subsubsection{Semantic Prior Validation.} 
ZOMG assumes that pretrained frame-level representations already encode meaningful temporal semantics, which can be further refined through instance-wise optimization. To validate this prior, we visualize the embeddings using t-SNE. As shown in Figure~\ref{fig:box} (bottom), frame embeddings form smooth trajectories within each sub-action, indicating coherent transitions and reflecting temporal semantics. This confirms that the pretrained space provides a strong structural prior, which our instance-specific optimization can exploit for fine-grained motion grounding.

\paragraph{Efficiency and Temporal Scaling.}
Table~\ref{tab:efficiency} shows ZOMG delivers strong grounding performance with minimal test-time cost. By optimizing only soft masks while keeping the encoder frozen, it achieves over 3× higher throughput than existing TTT methods, enabling practical deployment in latency-sensitive, annotation-free scenarios.
Fig.~\ref{fig:line} (a) further exhibits favorable \textit{temporal scaling}, where error steadily decreases with more optimization steps. This controllable trade-off between adaptation time and accuracy allows flexible use under different computational budgets.


\input{figs/box}

\subsection{Downstream Benefits}
To assess the broader utility of ZOMG, we augment motion retrieval datasets by generating fine-grained motion-text pairs grounded from sub-action queries, enriching compositional semantics beyond original annotations. We evaluate their impact on SOTA methods (MotionCLIP, TEMOS, TMR, MESM~\cite{shi2024modal}) under three standard protocols, including filtered and batch retrieval settings. For fair comparison, we include several augmentation baselines such as noise injection, temporal scaling, and motion concatenation, with all methods producing an equal number of samples. As shown in Table~\ref{tab:aug_retrival_humanml3d}, ZOMG consistently improves retrieval across models and settings, indicating that our grounded segments are both temporally precise and semantically discriminative. Full implementation details are in Appendix~A.5.

%% file: tables/grounding.tex
\definecolor{myblue}{RGB}{217,237,249}
\begin{table*}[htbp]
    \centering
\resizebox{0.85\textwidth}{!}{
\begin{tabular}{l|l|c|cccccc|c}
\toprule
\textbf{Dataset}         & \textbf{Method} & \textbf{TTT} & \textbf{AP@8$\uparrow$} & \textbf{AP@7$\uparrow$} & \textbf{AP@6$\uparrow$} & \textbf{AP@5$\uparrow$} & \textbf{AP@4$\uparrow$} & \textbf{AP@3$\uparrow$} & \textbf{mAP $\uparrow$ } \\
\midrule
\multirow{5}{*}{HumanML3D} &  MotionCLIP  & \XSolidBrush   & 0.00     &   2.13  &   4.61   &   12.06   &   17.85   &   24.29 & 10.16 \\
& TCAM & \XSolidBrush  & 1.06 & 5.14 & 13.12 & 19.92 & 31.09 & 36.58 & 17.82 \\
& FreeZAD & \XSolidBrush  & 7.68 & 14.17 & 21.09 & 27.24 & 34.81 & 42.76 & 24.63 \\ 
& T3AL & \Checkmark & 13.71 & 22.64 & 34.57 & 46.99 & 55.14 & 67.08 & 40.21 \\
& AdaZAD & \Checkmark & 15.11 & 24.39 & 38.42 & 48.18 & 56.87 & 67.63 & 41.77\\
&  \cellcolor{myblue}  ZOMG  & \cellcolor{myblue}\Checkmark &   \cellcolor{myblue}\textbf{28.25}   & \cellcolor{myblue}\textbf{37.53}    & \cellcolor{myblue}\textbf{51.42 }   &  \cellcolor{myblue}\textbf{53.90}    &  \cellcolor{myblue}\textbf{60.85}   & \cellcolor{myblue}\textbf{70.83}  & \cellcolor{myblue}\textbf{50.46}  \\

\midrule
\multirow{5}{*}{KIT-ML} &  MotionCLIP & \XSolidBrush & 0.00 & 1.58 & 3.72 & 9.62 & 17.31 & 23.93 &   9.36 \\
& TCAM  & \XSolidBrush & 3.85     &   7.54  &   12.85   &   21.10   &   28.46   &   35.23 & 18.17 \\
& FreeZAD & \XSolidBrush & 6.79 &9.34 & 14.39 & 22.62 & 27.43 & 37.50 & 19.68 \\
& T3AL &\Checkmark& 11.54 & 15.38 & 18.29 & 27.78 &  36.97 &  54.91 & 27.48  \\

& AdaZAD &\Checkmark& 13.68 & 17.91 & 24.57 & 32.14 & 39.49 &  56.88 & 30.78 \\
&   \cellcolor{myblue} ZOMG  & \cellcolor{myblue}\Checkmark &   \cellcolor{myblue}\textbf{22.01}   & \cellcolor{myblue}\textbf{25.64}    &  \cellcolor{myblue}\textbf{ 39.69} & \cellcolor{myblue}\textbf{43.74}  & \cellcolor{myblue}\textbf{48.59}       & \cellcolor{myblue}\textbf{62.73}  & \cellcolor{myblue}\textbf{40.40} \\
\midrule

\multirow{5}{*}{BABEL}  &  MotionCLIP& \XSolidBrush & 0.00 & 1.00 & 1.49 & 3.98 & 7.96 & 12.44 &   4.31 \\
& TCAM  & \XSolidBrush  & 0.00    &   1.00  &   2.49   &   4.98   &   11.44   &   17.91 & 6.30 \\
& FreeZAD & \XSolidBrush & 0.00  & 1.32 & 2.56  & 5.97 & 10.95 & 17.31 & 6.35 \\
& T3AL &\Checkmark& 0.23 & 1.86 & 3.41 & 8.09  & 16.38  &  29.57 & 9.92 \\
& AdaZAD &\Checkmark& \textbf{0.72} & \textbf{2.64} & 4.93 & 11.82 & 18.70 & 32.09 & 10.15 \\

& \cellcolor{myblue} ZOMG & \cellcolor{myblue}\Checkmark  & \cellcolor{myblue}0.50 &  \cellcolor{myblue}2.49 &  \cellcolor{myblue}\textbf{5.26}  & \cellcolor{myblue}\textbf{13.16} & \cellcolor{myblue}\textbf{23.03} & \cellcolor{myblue}\cellcolor{myblue} \textbf{36.18} & \cellcolor{myblue}\textbf{13.44}\\
\bottomrule

\end{tabular}
}
\caption{Motion grounding results on three public datasets. (TTT denotes Test-Time Training.)}
    \label{tab:grounding}
\end{table*}

%% file: figs/tsne.tex
\begin{figure}[tp]
    \centering
    \includegraphics[width=\linewidth]{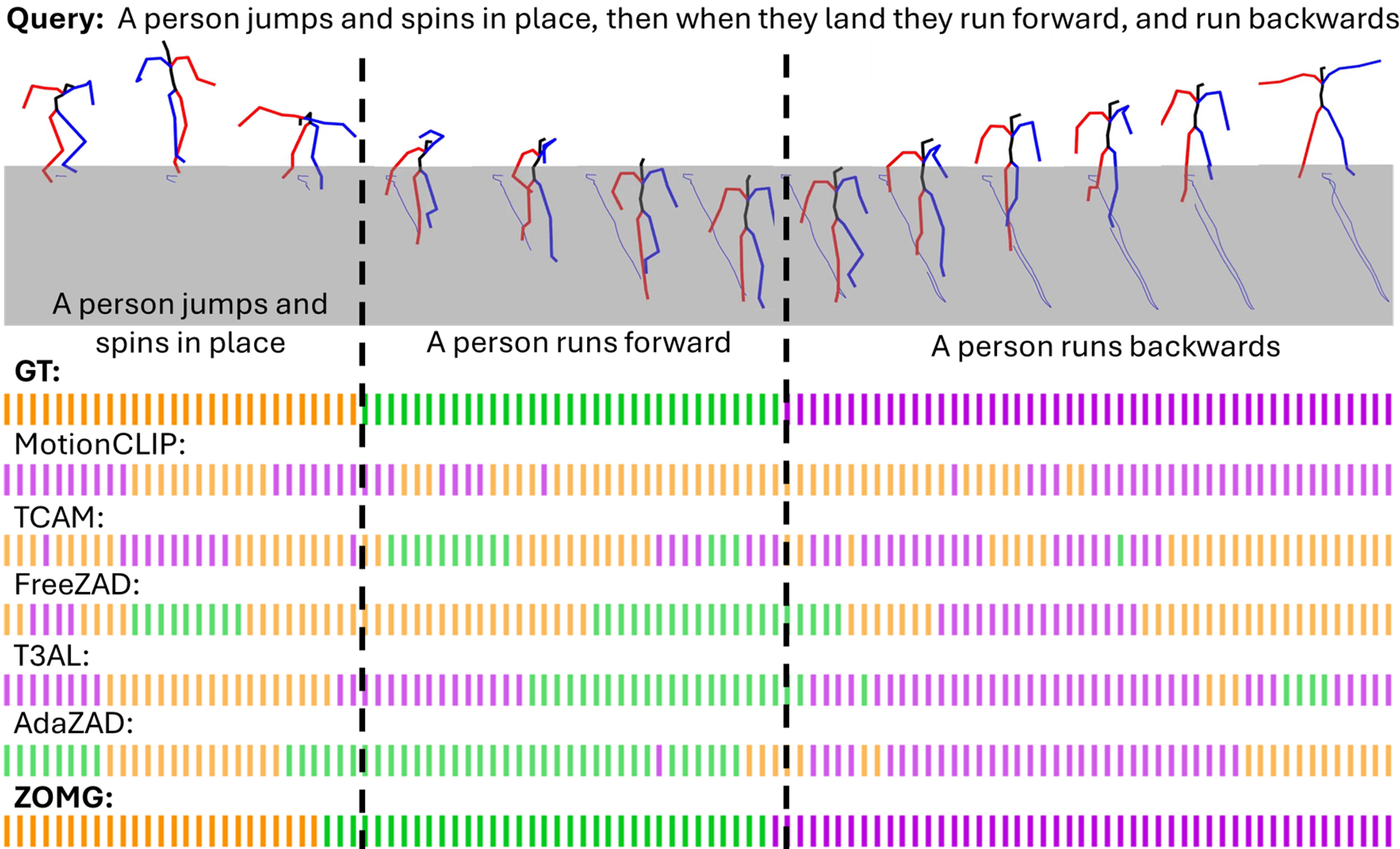}
    
    \caption{Motion grounding comparison in HumanML3D.}
    \label{fig:combo}
\end{figure}

%% file: figs/combo.tex
\begin{figure}[tp]
    \centering
    \includegraphics[width=\linewidth]{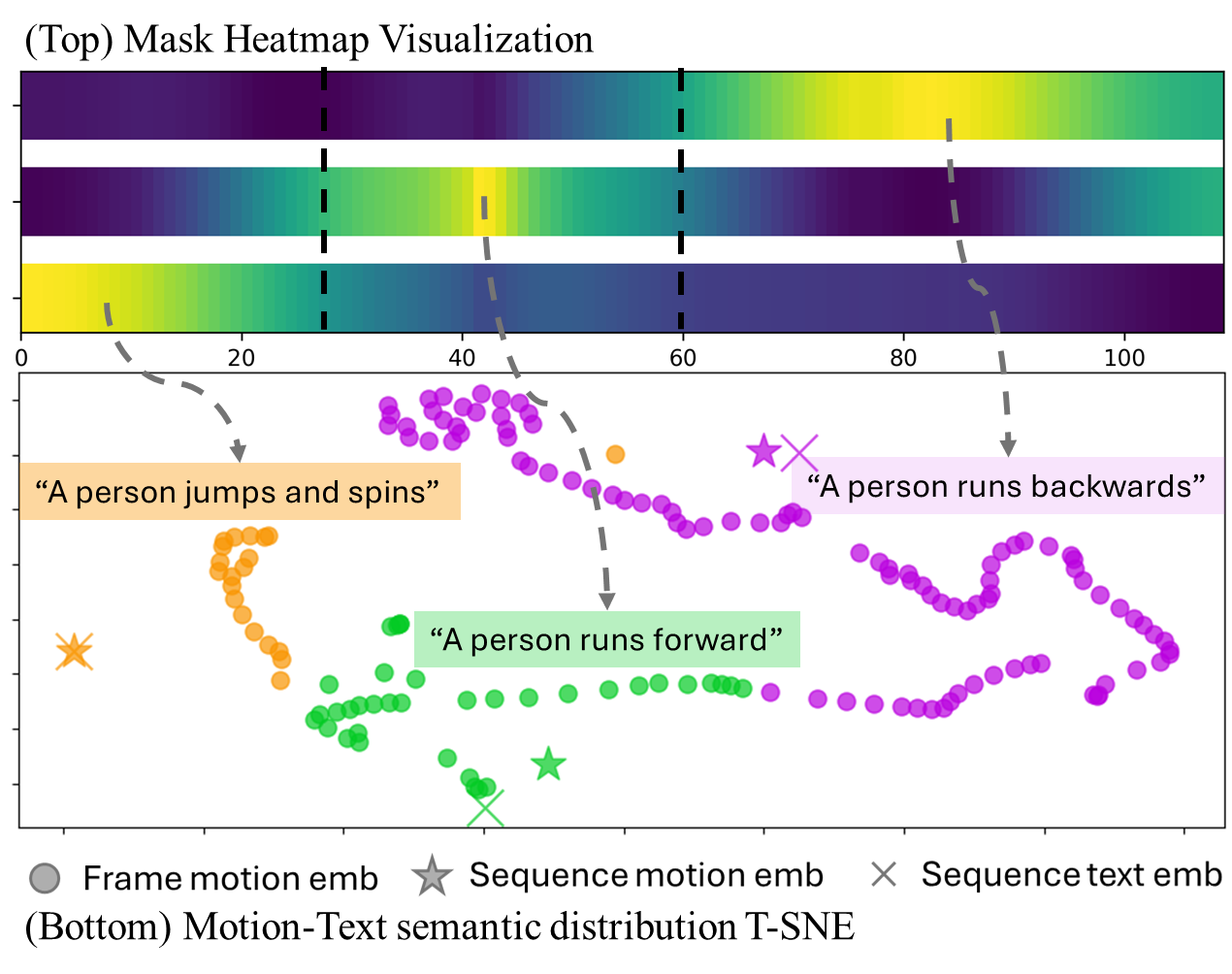}
    
    \caption{Analysis of ZOMG in (Top) mask heatmap, and (Bottom) T-SNE distribution.}
    \label{fig:box}
\end{figure}

%% file: figs/visual.tex
\begin{figure}[tp]
    \centering
    \includegraphics[width=\linewidth]{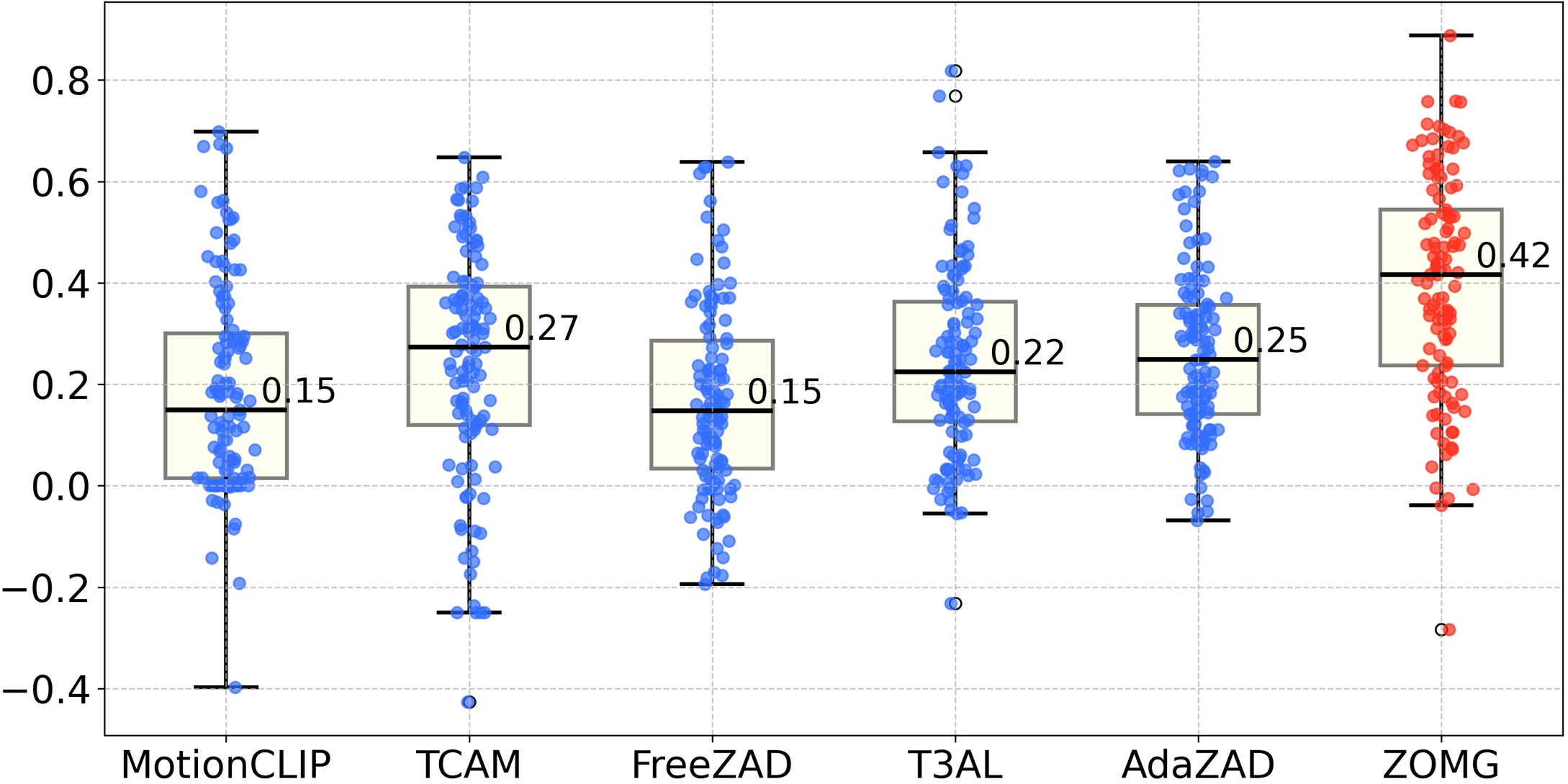}
    
    \caption{The boxplot comparison of semantic similarity for HumanML3D grounded motion-text pairs}
    \label{fig:visual}
\end{figure}

%% file: tables/ablation.tex
\definecolor{myblue}{RGB}{217,237,249}

\begin{table}[htbp]
    \centering
    \small
\begin{tabular}{c|c|ccc|cc|c}
\toprule
\multirow{2}{*}{\textbf{TTT}} & \multirow{2}{*}{\textbf{LSP}}  & \multicolumn{3}{c|}{\textbf{SMO}} & \multirow{2}{*}{\textbf{AP@7}}  & \multirow{2}{*}{\textbf{AP@3}} & \multirow{2}{*}{\textbf{mAP}} \\
\cmidrule{3-5}
& & \textbf{[$M$]} &  \textbf{$\hat{\mathcal{L}}_s$}   &  \textbf{$\hat{\mathcal{L}}_e$}    & & & \\
\midrule
\XSolidBrush  & \XSolidBrush  &\XSolidBrush & \XSolidBrush & \XSolidBrush & 0.91    &   14.33 & 5.86 \\
 \XSolidBrush & \Checkmark &\XSolidBrush & \XSolidBrush & \XSolidBrush & 2.13    &   24.29  & 10.16  \\
\Checkmark  & \Checkmark  &\XSolidBrush & \XSolidBrush & \XSolidBrush & 13.21    &   38.52  & 22.69  \\
\Checkmark & \Checkmark & \Checkmark & \XSolidBrush & \XSolidBrush & 21.17 & 62.90  & 38.63\\
\Checkmark  & \Checkmark &\Checkmark & \Checkmark & \XSolidBrush &  25.76 & 66.29 & 41.54 \\
\Checkmark  &\Checkmark & \Checkmark &  \XSolidBrush &\Checkmark &  33.51 &\textbf{ 71.20} & 47.28 \\

\rowcolor{myblue}\Checkmark &\Checkmark & \Checkmark & \Checkmark & \Checkmark & \textbf{37.53}   &  70.83    &  \textbf{50.46}   \\


\bottomrule

\end{tabular}
\caption{Ablation study of ZOMG on HumanML3D.}
    \label{tab:ablation}
\end{table}

%% file: tables/efficiency.tex
\begin{table}[h]
\centering
\resizebox{0.9\linewidth}{!}{
\begin{tabular}{l|ccc|c}
\toprule
\textbf{Method} & \textbf{Param.$\downarrow$} & \textbf{GFLOPs$\downarrow$} & \textbf{Samples/s$\uparrow$} & \textbf{mAP$\uparrow$} \\
\midrule
T3AL & 1.2 M & 2528.1  & 6.87 & 40.21\\
AdaZAD & 1.2 M & 2675.9 & 6.65 & 41.77\\
\rowcolor{myblue}ZOMG & \textbf{0.5 K}  & \textbf{302.5} & \textbf{23.25} & \textbf{50.46}  \\
\bottomrule
\end{tabular}
}
\caption{Computational costs during TTT for 100 steps.}
\label{tab:efficiency}
\end{table}

%% file: tables/aug_retrival_compress_hml3d.tex
\definecolor{myblue}{RGB}{217,237,249}

\begin{table*}[htbp]
    \centering
    \small
    \renewcommand{\arraystretch}{1.2} 
    \resizebox{0.9\textwidth}{!}{ 
    \begin{tabular}{lcl|cccc|cccc}
        \toprule
        \multirow{2}{*}{\textbf{Dataset}} & \multirow{2}{*}{\textbf{Protocol}} & \multirow{2}{*}{\textbf{Task}} & \multicolumn{4}{c}{\textbf{SOTA methods}} & \multicolumn{4}{c}{\textbf{Augmentation methods}} \\
        \cmidrule(lr){4-7} \cmidrule(lr){8-11}
        & & & \textbf{MotionCLIP} & \textbf{TEMOS} & \textbf{TMR} & \textbf{MESM} & \textbf{Noise} & \textbf{Scaling} & \textbf{Concat} & \textbf{ZOMG} \cellcolor{myblue} \\
        \midrule
        \multirow{6}{*}{\textbf{HumanML3D}} & \multirow{2}{*}{\textbf{A}} 
         & \textbf{T2M} & 16.00 & 13.15 & 16.32 & 19.20 & 17.05 & 17.15 & 17.15 & \textbf{19.38}\cellcolor{myblue} \\
        && \textbf{M2T} & 16.95 & 7.74 & 18.72 &\textbf{21.30} & 17.98 & 18.57 & 18.58 & 20.53\cellcolor{myblue}\\

        \cmidrule(lr){2-11}
        &\multirow{2}{*}{\textbf{B}} 
          & \textbf{T2M} & 22.49 & 12.36 & 22.75 & 26.29 & 23.87 & 23.96 & 23.90 & \textbf{27.01} \cellcolor{myblue}\\
        & & \textbf{M2T} & 21.69 & 9.96 & 23.25 & 25.09 & 23.19 & 23.61 & 23.71 & \textbf{25.52}\cellcolor{myblue}\\

        \cmidrule(lr){2-11}
        & \multirow{2}{*}{\textbf{C}} 
          & \textbf{T2M} & 86.90 & 62.05 & 84.42 & 86.22 & 85.92 & 86.27 & 85.89 & \textbf{87.58}\cellcolor{myblue}\\
        & & \textbf{M2T} & 87.12 & 62.71 & 84.50 & 86.15 & 86.18 & 86.72 & 85.93 & \textbf{87.56}\cellcolor{myblue}\\

    \midrule
    \multirow{6}{*}{\textbf{KIT-ML}} & \multirow{2}{*}{\textbf{A}} 
         & \textbf{T2M} & 22.49 & 19.54 & 22.00 & 23.76 & 23.19 & 23.61 & 23.86 & \textbf{24.03} \cellcolor{myblue}\\
        && \textbf{M2T} & 22.92 & 22.03 & 22.36 & 21.30 & 23.65 & 22.54 & 23.15 & \textbf{23.71}\cellcolor{myblue}\\

        \cmidrule(lr){2-11}
        &\multirow{2}{*}{\textbf{B}} 
          & \textbf{T2M}  & 40.22 & 34.48 & 41.52 & \textbf{43.04} & 39.66 & 39.15 & 40.07 & 42.04\cellcolor{myblue}\\
        & & \textbf{M2T} & 34.10 & 30.29 & 34.02 & 35.61 & 34.69 & 33.57 & 36.80 & \textbf{37.42}\cellcolor{myblue}\\

        \cmidrule(lr){2-11}
        & \multirow{2}{*}{\textbf{C}} 
          & \textbf{T2M} & 75.68 & 65.58 & 76.03 & 76.11 & 76.28 & 75.53 & 76.70 & \textbf{77.43}\cellcolor{myblue} \\
        & & \textbf{M2T} & 75.61 & 64.88 & 75.55 & 76.65  &75.94 & 75.80 & 75.80 & \textbf{76.99} \cellcolor{myblue}\\

        \bottomrule
    \end{tabular}
    }
    \caption{Text-to-motion retrieval performance comparison.}
    \label{tab:aug_retrival_humanml3d}
\end{table*}

%% file: figs/box.tex
\begin{figure}[tp]
    \centering
    \includegraphics[width=\linewidth]{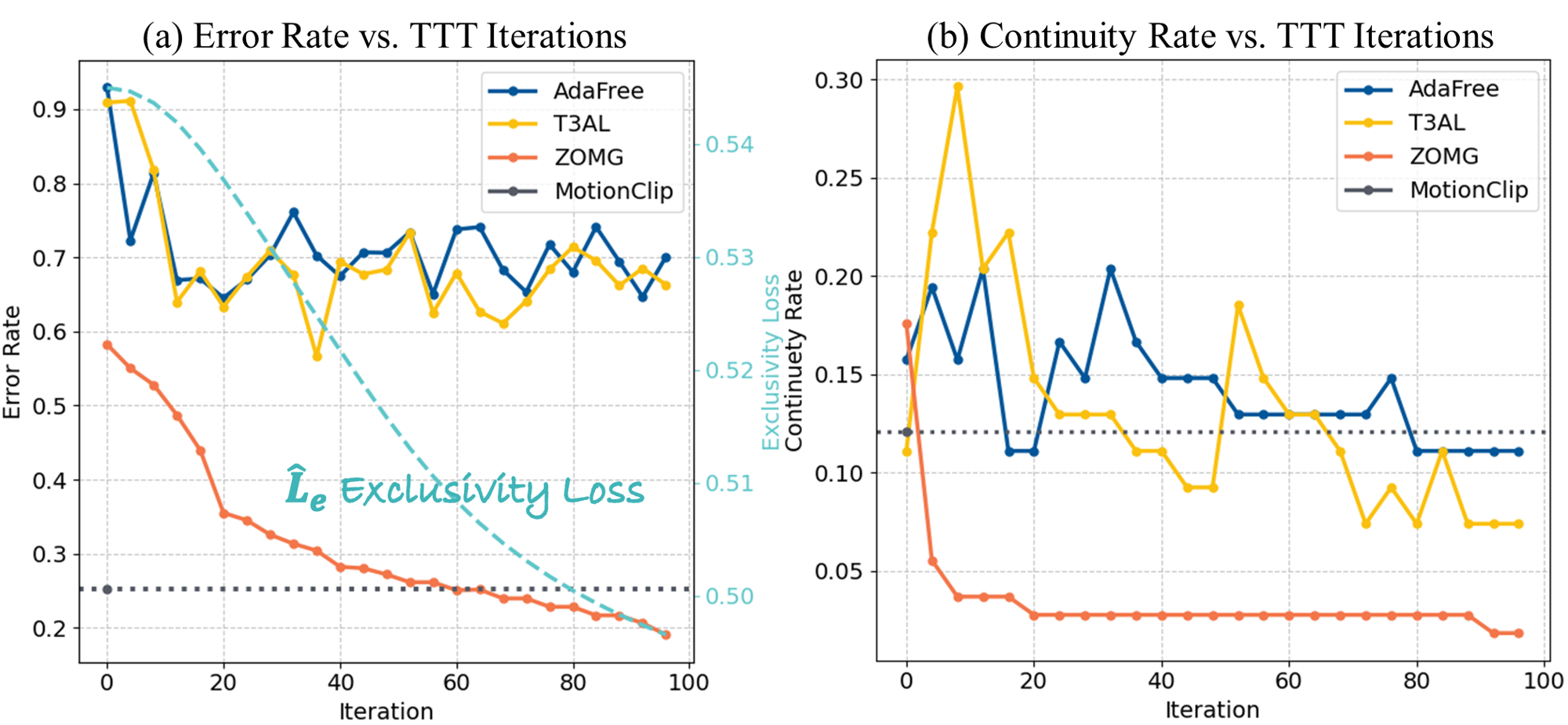}
    
    \caption{Comparison of (a) Mask constrain performance and (b) Motion-text semantic similarity on HumanML3D.}
    \label{fig:line}
\end{figure}

%% file: chapter/10_conclusion.tex
\section{Conclusion}
\label{sec:conclusion}

Our ZOMG demonstrates that accurate and open-vocabulary motion grounding can be achieved annotation-free, through lightweight test-time optimization without modifying pretrained models. 
Its high efficiency and strong performance across grounding and retrieval tasks highlight the potential of instance-adaptive inference for real-world deployment.
By uncovering compositional motion units in an unsupervised manner, ZOMG provides a scalable foundation for interpretable motion understanding and broad downstream transfer.

%% file: aaai2026.bib
@inproceedings{gao2017tall,
  title={Tall: Temporal activity localization via language query},
  author={Gao, Jiyang and Sun, Chen and Yang, Zhenheng and Nevatia, Ram},
  booktitle={Proceedings of the IEEE international conference on computer vision},
  pages={5267--5275},
  year={2017}
}

@inproceedings{zhang2020learning,
  title={Learning 2d temporal adjacent networks for moment localization with natural language},
  author={Zhang, Songyang and Peng, Houwen and Fu, Jianlong and Luo, Jiebo},
  booktitle={Proceedings of the AAAI conference on artificial intelligence},
  volume={34},
  number={07},
  pages={12870--12877},
  year={2020}
}

@inproceedings{liberatori2024test,
  title={Test-time zero-shot temporal action localization},
  author={Liberatori, Benedetta and Conti, Alessandro and Rota, Paolo and Wang, Yiming and Ricci, Elisa},
  booktitle={Proceedings of the IEEE/CVF conference on computer vision and pattern recognition},
  pages={18720--18729},
  year={2024}
}

@inproceedings{nag2022zero,
  title={Zero-shot temporal action detection via vision-language prompting},
  author={Nag, Sauradip and Zhu, Xiatian and Song, Yi-Zhe and Xiang, Tao},
  booktitle={European conference on computer vision},
  pages={681--697},
  year={2022},
  organization={Springer}
}

@article{han2025training,
  title={Training-Free Zero-Shot Temporal Action Detection with Vision-Language Models},
  author={Han, Chaolei and Wang, Hongsong and Kuang, Jidong and Zhang, Lei and Gui, Jie},
  journal={arXiv preprint arXiv:2501.13795},
  year={2025}
}

@inproceedings{nguyen2025multi,
  title={Multi-scale contrastive learning for video temporal grounding},
  author={Nguyen, Thong Thanh and Bin, Yi and Wu, Xiaobao and Hu, Zhiyuan and Nguyen, Cong-Duy T and Ng, See-Kiong and Luu, Anh Tuan},
  booktitle={Proceedings of the AAAI Conference on Artificial Intelligence},
  volume={39},
  number={6},
  pages={6227--6235},
  year={2025}
}

@inproceedings{gorti2022x,
  title={X-pool: Cross-modal language-video attention for text-video retrieval},
  author={Gorti, Satya Krishna and Vouitsis, No{\"e}l and Ma, Junwei and Golestan, Keyvan and Volkovs, Maksims and Garg, Animesh and Yu, Guangwei},
  booktitle={Proceedings of the IEEE/CVF conference on computer vision and pattern recognition},
  pages={5006--5015},
  year={2022}
}

@article{yu2023self,
  title={Self-chained image-language model for video localization and question answering},
  author={Yu, Shoubin and Cho, Jaemin and Yadav, Prateek and Bansal, Mohit},
  journal={Advances in Neural Information Processing Systems},
  volume={36},
  pages={76749--76771},
  year={2023}
}

@inproceedings{petrovich2022temos,
  title={Temos: Generating diverse human motions from textual descriptions},
  author={Petrovich, Mathis and Black, Michael J and Varol, G{\"u}l},
  booktitle={European Conference on Computer Vision},
  pages={480--497},
  year={2022},
  organization={Springer}
}

@inproceedings{petrovich2023tmr,
  title={Tmr: Text-to-motion retrieval using contrastive 3d human motion synthesis},
  author={Petrovich, Mathis and Black, Michael J and Varol, G{\"u}l},
  booktitle={Proceedings of the IEEE/CVF International Conference on Computer Vision},
  pages={9488--9497},
  year={2023}
}

@inproceedings{athanasiou2022teach,
  title={Teach: Temporal action composition for 3d humans},
  author={Athanasiou, Nikos and Petrovich, Mathis and Black, Michael J and Varol, G{\"u}l},
  booktitle={2022 International Conference on 3D Vision (3DV)},
  pages={414--423},
  year={2022},
  organization={IEEE}
}

@inproceedings{tevet2022motionclip,
  title={Motionclip: Exposing human motion generation to clip space},
  author={Tevet, Guy and Gordon, Brian and Hertz, Amir and Bermano, Amit H and Cohen-Or, Daniel},
  booktitle={European Conference on Computer Vision},
  pages={358--374},
  year={2022},
  organization={Springer}
}

@article{lin2023motion,
  title={Motion-x: A large-scale 3d expressive whole-body human motion dataset},
  author={Lin, Jing and Zeng, Ailing and Lu, Shunlin and Cai, Yuanhao and Zhang, Ruimao and Wang, Haoqian and Zhang, Lei},
  journal={Advances in Neural Information Processing Systems},
  volume={36},
  pages={25268--25280},
  year={2023}
}

@article{wang2023actionclip,
  title={Actionclip: Adapting language-image pretrained models for video action recognition},
  author={Wang, Mengmeng and Xing, Jiazheng and Mei, Jianbiao and Liu, Yong and Jiang, Yunliang},
  journal={IEEE Transactions on Neural Networks and Learning Systems},
  year={2023},
  publisher={IEEE}
}

@inproceedings{kong2019mmact,
  title={Mmact: A large-scale dataset for cross modal human action understanding},
  author={Kong, Quan and Wu, Ziming and Deng, Ziwei and Klinkigt, Martin and Tong, Bin and Murakami, Tomokazu},
  booktitle={Proceedings of the IEEE/CVF International Conference on Computer Vision},
  pages={8658--8667},
  year={2019}
}

@inproceedings{sun2020test,
  title={Test-time training with self-supervision for generalization under distribution shifts},
  author={Sun, Yu and Wang, Xiaolong and Liu, Zhuang and Miller, John and Efros, Alexei and Hardt, Moritz},
  booktitle={International conference on machine learning},
  pages={9229--9248},
  year={2020},
  organization={PMLR}
}

@article{liu2016large,
  title={Large-margin softmax loss for convolutional neural networks},
  author={Liu, Weiyang and Wen, Yandong and Yu, Zhiding and Yang, Meng},
  journal={arXiv preprint arXiv:1612.02295},
  year={2016}
}

@inproceedings{wortsman2022robust,
  title={Robust fine-tuning of zero-shot models},
  author={Wortsman, Mitchell and Ilharco, Gabriel and Kim, Jong Wook and Li, Mike and Kornblith, Simon and Roelofs, Rebecca and Lopes, Raphael Gontijo and Hajishirzi, Hannaneh and Farhadi, Ali and Namkoong, Hongseok and others},
  booktitle={Proceedings of the IEEE/CVF conference on computer vision and pattern recognition},
  pages={7959--7971},
  year={2022}
}

@inproceedings{lee2019self,
  title={Self-attention graph pooling},
  author={Lee, Junhyun and Lee, Inyeop and Kang, Jaewoo},
  booktitle={International conference on machine learning},
  pages={3734--3743},
  year={2019},
  organization={pmlr}
}

@article{wang2020k,
  title={K-adapter: Infusing knowledge into pre-trained models with adapters},
  author={Wang, Ruize and Tang, Duyu and Duan, Nan and Wei, Zhongyu and Huang, Xuanjing and Cao, Guihong and Jiang, Daxin and Zhou, Ming and others},
  journal={arXiv preprint arXiv:2002.01808},
  year={2020}
}

@inproceedings{radford2021learning,
  title={Learning transferable visual models from natural language supervision},
  author={Radford, Alec and Kim, Jong Wook and Hallacy, Chris and Ramesh, Aditya and Goh, Gabriel and Agarwal, Sandhini and Sastry, Girish and Askell, Amanda and Mishkin, Pamela and Clark, Jack and others},
  booktitle={International conference on machine learning},
  pages={8748--8763},
  year={2021},
  organization={PmLR}
}

@article{wang2023glanet,
  title={GLANet: temporal knowledge graph completion based on global and local information-aware network},
  author={Wang, Jingbin and Lin, Xinyu and Huang, Hao and Ke, Xifan and Wu, Renfei and You, Changkai and Guo, Kun},
  journal={Applied Intelligence},
  volume={53},
  number={16},
  pages={19285--19301},
  year={2023},
  publisher={Springer}
}

@article{liu2021ttt++,
  title={Ttt++: When does self-supervised test-time training fail or thrive?},
  author={Liu, Yuejiang and Kothari, Parth and Van Delft, Bastien and Bellot-Gurlet, Baptiste and Mordan, Taylor and Alahi, Alexandre},
  journal={Advances in Neural Information Processing Systems},
  volume={34},
  pages={21808--21820},
  year={2021}
}

@article{fitzgerald2018large,
  title={Large-scale QA-SRL parsing},
  author={FitzGerald, Nicholas and Michael, Julian and He, Luheng and Zettlemoyer, Luke},
  journal={arXiv preprint arXiv:1805.05377},
  year={2018}
}

@article{zhao2023survey,
  title={A survey of large language models},
  author={Zhao, Wayne Xin and Zhou, Kun and Li, Junyi and Tang, Tianyi and Wang, Xiaolei and Hou, Yupeng and Min, Yingqian and Zhang, Beichen and Zhang, Junjie and Dong, Zican and others},
  journal={arXiv preprint arXiv:2303.18223},
  volume={1},
  number={2},
  year={2023}
}

@article{wang2024text,
  title={Text-controlled motion mamba: text-instructed temporal grounding of human motion},
  author={Wang, Xinghan and Kang, Zixi and Mu, Yadong},
  journal={arXiv preprint arXiv:2404.11375},
  year={2024}
}

@article{zhang2023finemogen,
  title={Finemogen: Fine-grained spatio-temporal motion generation and editing},
  author={Zhang, Mingyuan and Li, Huirong and Cai, Zhongang and Ren, Jiawei and Yang, Lei and Liu, Ziwei},
  journal={Advances in Neural Information Processing Systems},
  volume={36},
  pages={13981--13992},
  year={2023}
}

@article{li2024motion,
  title={Motion generation from fine-grained textual descriptions},
  author={Li, Kunhang and Feng, Yansong},
  journal={arXiv preprint arXiv:2403.13518},
  year={2024}
}

@inproceedings{punnakkal2021babel,
  title={BABEL: Bodies, action and behavior with english labels},
  author={Punnakkal, Abhinanda R and Chandrasekaran, Arjun and Athanasiou, Nikos and Quiros-Ramirez, Alejandra and Black, Michael J},
  booktitle={Proceedings of the IEEE/CVF Conference on Computer Vision and Pattern Recognition},
  pages={722--731},
  year={2021}
}

@inproceedings{chao2018rethinking,
  title={Rethinking the faster r-cnn architecture for temporal action localization},
  author={Chao, Yu-Wei and Vijayanarasimhan, Sudheendra and Seybold, Bryan and Ross, David A and Deng, Jia and Sukthankar, Rahul},
  booktitle={Proceedings of the IEEE conference on computer vision and pattern recognition},
  pages={1130--1139},
  year={2018}
}

@inproceedings{zhao2017temporal,
  title={Temporal action detection with structured segment networks},
  author={Zhao, Yue and Xiong, Yuanjun and Wang, Limin and Wu, Zhirong and Tang, Xiaoou and Lin, Dahua},
  booktitle={Proceedings of the IEEE international conference on computer vision},
  pages={2914--2923},
  year={2017}
}

@inproceedings{yan2023unloc,
  title={Unloc: A unified framework for video localization tasks},
  author={Yan, Shen and Xiong, Xuehan and Nagrani, Arsha and Arnab, Anurag and Wang, Zhonghao and Ge, Weina and Ross, David and Schmid, Cordelia},
  booktitle={Proceedings of the IEEE/CVF International Conference on Computer Vision},
  pages={13623--13633},
  year={2023}
}

@article{bordes2024introduction,
  title={An introduction to vision-language modeling},
  author={Bordes, Florian and Pang, Richard Yuanzhe and Ajay, Anurag and Li, Alexander C and Bardes, Adrien and Petryk, Suzanne and Ma{\~n}as, Oscar and Lin, Zhiqiu and Mahmoud, Anas and Jayaraman, Bargav and others},
  journal={arXiv preprint arXiv:2405.17247},
  year={2024}
}

@article{xu2021vlm,
  title={Vlm: Task-agnostic video-language model pre-training for video understanding},
  author={Xu, Hu and Ghosh, Gargi and Huang, Po-Yao and Arora, Prahal and Aminzadeh, Masoumeh and Feichtenhofer, Christoph and Metze, Florian and Zettlemoyer, Luke},
  journal={arXiv preprint arXiv:2105.09996},
  year={2021}
}

@article{lu2023humantomato,
  title={Humantomato: Text-aligned whole-body motion generation},
  author={Lu, Shunlin and Chen, Ling-Hao and Zeng, Ailing and Lin, Jing and Zhang, Ruimao and Zhang, Lei and Shum, Heung-Yeung},
  journal={arXiv preprint arXiv:2310.12978},
  year={2023}
}

@inproceedings{guo2022generating,
  title={Generating diverse and natural 3d human motions from text},
  author={Guo, Chuan and Zou, Shihao and Zuo, Xinxin and Wang, Sen and Ji, Wei and Li, Xingyu and Cheng, Li},
  booktitle={Proceedings of the IEEE/CVF conference on computer vision and pattern recognition},
  pages={5152--5161},
  year={2022}
}

@article{plappert2016kit,
  title={The kit motion-language dataset},
  author={Plappert, Matthias and Mandery, Christian and Asfour, Tamim},
  journal={Big data},
  volume={4},
  number={4},
  pages={236--252},
  year={2016},
  publisher={Mary Ann Liebert, Inc. 140 Huguenot Street, 3rd Floor New Rochelle, NY 10801 USA}
}

@inproceedings{liu2022exploring,
  title={Exploring motion and appearance information for temporal sentence grounding},
  author={Liu, Daizong and Qu, Xiaoye and Zhou, Pan and Liu, Yang},
  booktitle={Proceedings of the AAAI Conference on Artificial Intelligence},
  volume={36},
  number={2},
  pages={1674--1682},
  year={2022}
}

@article{wu2022rule,
  title={Rule-based information extraction for mechanical-electrical-plumbing-specific semantic web},
  author={Wu, Lang-Tao and Lin, Jia-Rui and Leng, Shuo and Li, Jiu-Lin and Hu, Zhen-Zhong},
  journal={Automation in Construction},
  volume={135},
  pages={104108},
  year={2022},
  publisher={Elsevier}
}

@article{yang2025qwen3,
  title={Qwen3 Technical Report},
  author={Yang, An and Li, Anfeng and Yang, Baosong and Zhang, Beichen and Hui, Binyuan and Zheng, Bo and Yu, Bowen and Gao, Chang and Huang, Chengen and Lv, Chenxu and others},
  journal={arXiv preprint arXiv:2505.09388},
  year={2025}
}

@inproceedings{belharbi2023tcam,
  title={Tcam: Temporal class activation maps for object localization in weakly-labeled unconstrained videos},
  author={Belharbi, Soufiane and Ben Ayed, Ismail and McCaffrey, Luke and Granger, Eric},
  booktitle={Proceedings of the IEEE/CVF Winter Conference on Applications of Computer Vision},
  pages={137--146},
  year={2023}
}

@inproceedings{zeng2025light,
  title={Light-t2m: A lightweight and fast model for text-to-motion generation},
  author={Zeng, Ling-An and Huang, Guohong and Wu, Gaojie and Zheng, Wei-Shi},
  booktitle={Proceedings of the AAAI Conference on Artificial Intelligence},
  volume={39},
  number={9},
  pages={9797--9805},
  year={2025}
}

@inproceedings{yang2022tubedetr,
  title={Tubedetr: Spatio-temporal video grounding with transformers},
  author={Yang, Antoine and Miech, Antoine and Sivic, Josef and Laptev, Ivan and Schmid, Cordelia},
  booktitle={Proceedings of the IEEE/CVF Conference on Computer Vision and Pattern Recognition},
  pages={16442--16453},
  year={2022}
}

@article{chen2021end,
  title={End-to-end multi-modal video temporal grounding},
  author={Chen, Yi-Wen and Tsai, Yi-Hsuan and Yang, Ming-Hsuan},
  journal={Advances in Neural Information Processing Systems},
  volume={34},
  pages={28442--28453},
  year={2021}
}

@inproceedings{shi2024modal,
  title={Modal-Enhanced Semantic Modeling for Fine-Grained 3D Human Motion Retrieval},
  author={Shi, Haoyu and Zhang, Huaiwen},
  booktitle={Proceedings of the 32nd ACM International Conference on Multimedia},
  pages={10114--10123},
  year={2024}
}

@inproceedings{zhou2024avatargpt,
  title={Avatargpt: All-in-one framework for motion understanding planning generation and beyond},
  author={Zhou, Zixiang and Wan, Yu and Wang, Baoyuan},
  booktitle={Proceedings of the IEEE/CVF Conference on Computer Vision and Pattern Recognition},
  pages={1357--1366},
  year={2024}
}

@inproceedings{xue2025shotvl,
  title={ShotVL: Human-Centric Highlight Frame Retrieval via Language Queries},
  author={Xue, Wangyu and Qian, Chen and Wu, Jiayi and Zhou, Yang and Liu, Wentao and Ren, Ju and Fan, Siming and Zhang, Yaoxue},
  booktitle={Proceedings of the AAAI Conference on Artificial Intelligence},
  volume={39},
  number={9},
  pages={9050--9058},
  year={2025}
}

@article{qian2025beyond,
  title={Beyond the Next Token: Towards Prompt-Robust Zero-Shot Classification via Efficient Multi-Token Prediction},
  author={Qian, Junlang and Zhu, Zixiao and Zhou, Hanzhang and Feng, Zijian and Zhai, Zepeng and Mao, Kezhi},
  journal={arXiv preprint arXiv:2504.03159},
  year={2025}
}

@article{zhou2023metafi++,
  title={MetaFi++: WiFi-enabled transformer-based human pose estimation for metaverse avatar simulation},
  author={Zhou, Yunjiao and Huang, He and Yuan, Shenghai and Zou, Han and Xie, Lihua and Yang, Jianfei},
  journal={IEEE Internet of Things Journal},
  volume={10},
  number={16},
  pages={14128--14136},
  year={2023},
  publisher={IEEE}
}

@article{zhou2024adapose,
  title={Adapose: Towards cross-site device-free human pose estimation with commodity wifi},
  author={Zhou, Yunjiao and Yang, Jianfei and Huang, He and Xie, Lihua},
  journal={IEEE Internet of Things Journal},
  year={2024},
  publisher={IEEE}
}

@inproceedings{yang2022metafi,
  title={MetaFi: Device-free pose estimation via commodity WiFi for metaverse avatar simulation},
  author={Yang, Jianfei and Zhou, Yunjiao and Huang, He and Zou, Han and Xie, Lihua},
  booktitle={2022 IEEE 8th World Forum on Internet of Things (WF-IoT)},
  pages={1--6},
  year={2022},
  organization={IEEE}
}

@article{zhou2023tent,
  title={Tent: Connect language models with iot sensors for zero-shot activity recognition},
  author={Zhou, Yunjiao and Yang, Jianfei and Zou, Han and Xie, Lihua},
  journal={arXiv preprint arXiv:2311.08245},
  year={2023}
}
